\documentclass[10pt,twocolumn,letterpaper]{article}

\usepackage{cvpr}              

\definecolor{cvprblue}{rgb}{0.21,0.49,0.74}
\usepackage[pagebackref,breaklinks,colorlinks,allcolors=cvprblue]{hyperref}
\usepackage{bm}
\usepackage{bbding}
\usepackage{graphicx}
\usepackage{tabularx}
\usepackage{booktabs}
\usepackage{algorithm}
\usepackage{algorithmic}
\def\paperID{3439} 
\def\confName{CVPR}
\def\confYear{2025}

\title{Bayesian Test-Time Adaptation for Vision-Language Models}

\author{
Lihua Zhou$^{1}$,
Mao Ye$^{2}$,
Shuaifeng Li$^{2}$,
Nianxin Li$^{2}$,
Xiatian Zhu$^{3}$,
Lei Deng$^{4}$,\\
Hongbin Liu$^{1,5}$,
Zhen Lei$^{1,5,6,7}$\thanks{Corresponding author}
\\
$^{1}$ CAIR, HKSIS, CAS,
$^{2}$ UESTC,
$^{3}$ University of Surrey,
$^{4}$ Shenzhen University,\\
$^{5}$ MAIS, Institute of Automation, CAS,
$^{6}$ SAI, UCAS,
$^{7}$ M.U.S.T
\\
}

\begin{document}
\maketitle
\begin{abstract}
Test-time adaptation with pre-trained vision-language models, such as CLIP, aims to adapt the model to new, potentially out-of-distribution test data.  Existing methods calculate the similarity between visual embedding and learnable class embeddings, which are initialized by text embeddings, for zero-shot image classification. In this work, we first analyze this process based on Bayes theorem, and observe that the core factors influencing the final prediction are the likelihood and the prior. However, existing methods essentially focus on adapting class embeddings to adapt likelihood, but they often ignore the importance of prior. To address this gap, we propose a novel approach, \textbf{B}ayesian \textbf{C}lass \textbf{A}daptation (BCA), which in addition to continuously updating class embeddings to adapt likelihood, also uses the posterior of incoming samples to continuously update the prior for each class embedding. This dual updating mechanism allows the model to better adapt to distribution shifts and achieve higher prediction accuracy. Our method not only surpasses existing approaches in terms of performance metrics but also maintains superior inference rates and memory usage, making it highly efficient and practical for real-world applications. 
\end{abstract}

\section{Introduction}\label{sec:intro}

Recent advances in visual tasks have seen a growing preference for using pre-trained vision-language models like CLIP \cite{radford2021learning,jia2021scaling,yang2022vision,lavoiemodeling}, which leverage large-scale paired image-text data enables it to learn rich, multi-modal representations that generalize well to a variety of downstream tasks \cite{li2024cloud}. However, when deploying CLIP in real-world environments, the distribution discrepancy between the pre-trained data and the environment data can lead to performance degradation \cite{liang2024comprehensive,Xiong_2023_ICCV,li2022source,Zhou_2023_ICCV,zhou2023adaptive}. To address this issue in real-time, test-time adaptation (TTA), which aims adapt the pre-trained model to new, potentially out-of-distribution test data during the inference phase, have been applied to the CLIP \cite{karmanov2024efficient,han2024dota,feng2023diverse,shu2022test}.

\begin{figure}[t]
  \begin{center}
  \includegraphics[width=\linewidth]{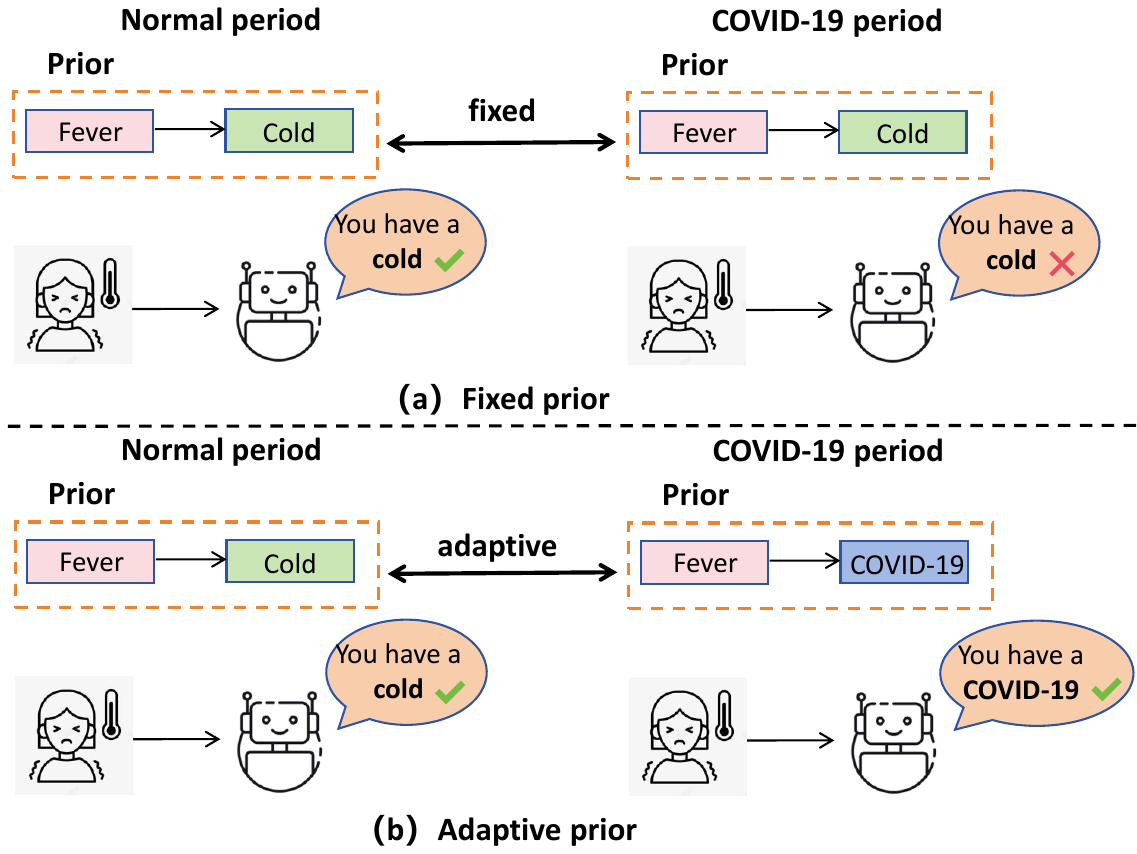}
  \end{center}
     \caption{Fixed Prior \textit{vs.} Adaptive Prior: Comparison of Diagnosis Outcomes. In the fixed prior scenario, patients with fever are consistently diagnosed with the common cold, regardless of whether it is a normal period or a COVID-19 period. In contrast, the adaptive prior scenario adjusts the diagnosis based on the current context. During normal periods, patients with fever are diagnosed with the common cold, while during the COVID-19 period, they are more likely to be diagnosed with COVID-19. This demonstrates the importance of performing prior adaptation in different environments.
}
  \label{fig:1}
  \vspace{-0.5cm}
\end{figure}

Existing TTA methods based on CLIP can be roughly divided into two categories. 
Early methods \cite{feng2023diverse, shu2022test} fine-tune the text prompts by minimizing the entropy of prediction to generate new class embeddings, which requires backpropagation and can be time-consuming, and then calculate the similarity between visual embedding and these new class embeddings for zero-shot classification. 
In fact, this approach is not in line with the TTA problem setting, which emphasizes real-time adaptation. 
Therefore, recent methods \cite{karmanov2024efficient, han2024dota} have begun to use memory to continuously store visual embeddings to increase the number of class embeddings or employ statistical methods to update class embeddings with incoming visual embedding, and then perform classification with these updated class embeddings. These approaches can perform real-time test-time adaptation, aligning more closely with the TTA problem setting.

Essentially, current test-time adaptation methods based on CLIP generate initial class embeddings $\bm{U}=\{\bm{\mu}_m\}_{m=1}^M$ by encoding text prompts and dynamically 
update these class embeddings $\bm{U}$ to update the logits (or probability) $P(\bm{x}|\bm{U})$, which is also known as the likelihood, to better adapt to the test environment distribution during the test phase, thereby making a better prediction $P(Y|\bm{x})$, which is also called the posterior.
In this work, we leverage the Bayes theorem to analyze and observe that the posterior $P(Y|\bm{x})$ is influenced not only by the likelihood $P(\bm{x}|\bm{U})$ but also by the prior $P(Y|\bm{U})$, which reflects initial belief about the categories $Y$ given the class embeddings $\bm{U}$. However, existing methods \cite{karmanov2024efficient,han2024dota,feng2023diverse,shu2022test} all ignore this point and use the fixed prior from pre-trained CLIP, which inevitably leads to suboptimal solutions when the test data distribution significantly deviates from the pre-trained data distribution. We also give a related example in Figure \ref{fig:1}. Please note that we focus on the conditional prior $P({Y}|\bm{U})$ rather than $P({Y})$. Unless otherwise specified, the prior we refer to in this work is $P({Y}|\bm{U})$. 



In this work, we propose \textbf{B}ayesian \textbf{C}lass \textbf{A}daptation (BCA), which not only considers only consider the likelihood adaptation as in previous methods, but also incorporate the prior adaptation. 
When a test image arrives, its visual embedding is computed using the visual encoder of CLIP, and the probability of belonging to each class embedding is calculated based on the current likelihood. For likelihood adaptation, the most similar class embedding is selected and updated using a statistical method by recalculating the mean of the selected class embedding with the current visual embedding. For prior adaptation, the probability of belonging to each class embedding are combined with the current prior to compute the posterior for each category. The category with the highest posterior probability is selected as the most likely category for the image. 
The prior of the selected class embedding is updated by recalculating the mean with the posterior.
This dynamic update mechanism helps the model adapt to the test distribution, thereby enhancing the robustness and accuracy of zero-shot image classification in real-world applications. Despite its simplicity, the method ensures high inference rates and low memory usage, making it highly suitable for TTA scenarios.
Our contributions are as follows:
\begin{enumerate}
\item \textbf{Comprehensive Analysis:} We leverage the Bayes theorem to analyze the process of CLIP performing zero-shot image classification, observing that the key of making better prediction lies in considering both the likelihood adaptation and the prior adaptation. We find that existing methods often ignore the prior adaptation. By addressing this issue, we propose a more general method.

\item \textbf{Effective and Fast Method:} We propose a efficient method that not only achieves good performance on Out-of-Distribution (OOD) and Cross Domain benchmarks, but also maintains high inference speed and low memory usage, making it highly suitable for TTA scenarios.
\end{enumerate}

\vspace{-0.3cm}
\section{Related Work}\label{sec:related_work}

\noindent\textbf{Vision-language models.}
Most visual recognition studies rely on training a separate deep neural network for each task, which is laborious \cite{he2016deep}. Recently, Vision-Language Models (VLMs) have emerged, learning rich vision-language correlations from web-scale image-text pairs and enabling zero-shot predictions on various visual recognition tasks with a single model \cite{radford2021learning,jia2021scaling,yaofilip,chenpali}.
CLIP \cite{radford2021learning} is the pioneering work that uses contrastive learning to align image and text embeddings.
ALIGN \cite{jia2021scaling} scales up the training of vision-language models by leveraging a dataset of over one billion noisy image alt-text pairs.
FILIP \cite{yaofilip} models the fine-grained semantic alignment through a novel cross-modal late interaction mechanism in the contrastive loss.
PaLI \cite{chenpali} performs joint scaling on both the language and vision components for a wide range of parameters.

\noindent\textbf{Transfer learning with vision-language models.}
Since pre-trained VLMs are generally trained to learn task-agnostic concepts, their performance can degrade when applied to downstream tasks due to the task-specific styles of images and text. To address this issue, transfer learning has been introduced to further adapt them to downstream tasks. From a methodological perspective, these methods can be roughly divided into two categories: prompt tuning \cite{zhou2022learning,zhou2022conditional,bahng2022exploring,rong2023retrieval,zang2022unified,shen2024multitask} and feature adapter \cite{gao2024clip,zhang2021tip,pantazis2022svl}.

For prompt tuning, it fits the downstream tasks by finding the optimal prompts. 
CoOp \cite{zhou2022learning} propose Context Optimization to learn a specific set of context tokens for each class, while CoCoOp \cite{zhou2022conditional} further propose conditional Context Optimization to generate a specific prompt for each image.
Unlike these two methods perform text prompt tuning, VP \cite{bahng2022exploring} uses a learnable perturbations on images to perform visual prompt tuning. RePrompt \cite{rong2023retrieval} proposes retrieval-enhanced visual prompt learning to cache the knowledge of downstream tasks.
Further, UPT \cite{zang2022unified} tunes both the visual and text prompt under the few-shot learning and domain generalization settings. MVLPT \cite{shen2024multitask} incorporates cross-task knowledge to enable information sharing for visual and text prompt tuning.
For feature adapter, it fine-tunes VLMs to align both text embedding and visual embedding with a light-weight adapter. 
CLIP-Adapter \cite{gao2024clip} adopts an additional bottleneck layer to learn new features and performs residual-style feature blending with the original pretrained features.
Tip-Adapter \cite{zhang2021tip}  creates the weights by a key-value cache model constructed from the few-shot training set.
SVL-Adapter \cite{pantazis2022svl} introduces a new encoder which is trained using self-supervision on the target dataset to combine the strengths of both vision-language pretraining and self-supervised learning.

\noindent\textbf{{Test-time adaptation with vision-language models.}} Test-time adaptation aims to adapt pre-trained model to test data, which may have the distribution discrepancy with the training data, during the test time. Early TTA methods often applied entropy minimization \cite{mirza2022norm,zhaodelta}, batch normalization calibration \cite{niutowards,wangtent}, pseudo-labeling \cite{iwasawa2021test,jangtest} and consistency regularization \cite{wang2022continual,niu2022efficient} to improve the performance.

Recently, with the increasing popularity of VLMs, many TTA works have also been applied to CLIP. 
TPT \cite{shu2022test} optimizes the text prompt to encourage consistent predictions across augmented views by minimizing the marginal entropy. 
DiffTPT \cite{feng2023diverse} leverages pre-trained diffusion models to generate diverse and informative augmented data to improve TPT. 
C-TPT \cite{yoonc} jointly optimizes the prompt to achieve better calibration by maximizing average text feature dispersion during test time.
Unlike these methods that require backpropagation during the test time,
TDA \cite{karmanov2024efficient} adapts to test data gradually via progressive pseudo label refinement with a lightweight key-value cache.
DOTA \cite{han2024dota} continually estimates the distribution of test samples to adapt the model work well in test environment.
MTA \cite{zanella2024test} manages augmented views by optimizing inlierness variables to generate final visual embeddings in a training-free manner.
PromptAlign \cite{samadhalign} explicitly aligns the test sample statistics with that of the source data distribution through token distribution alignment to optimizes the text prompt.
HisTPT \cite{zhang2024historical} introduces three types of knowledge banks to build up comprehensive memorization, mitigating the knowledge forgetting.
Zero \cite{farina2024frustratingly} sets the Softmax temperature to zero to converting probability distributions into one-hot encodings for test-time adaptation.

\begin{figure*}[t]
  \begin{center}
  \includegraphics[width=0.9\linewidth]{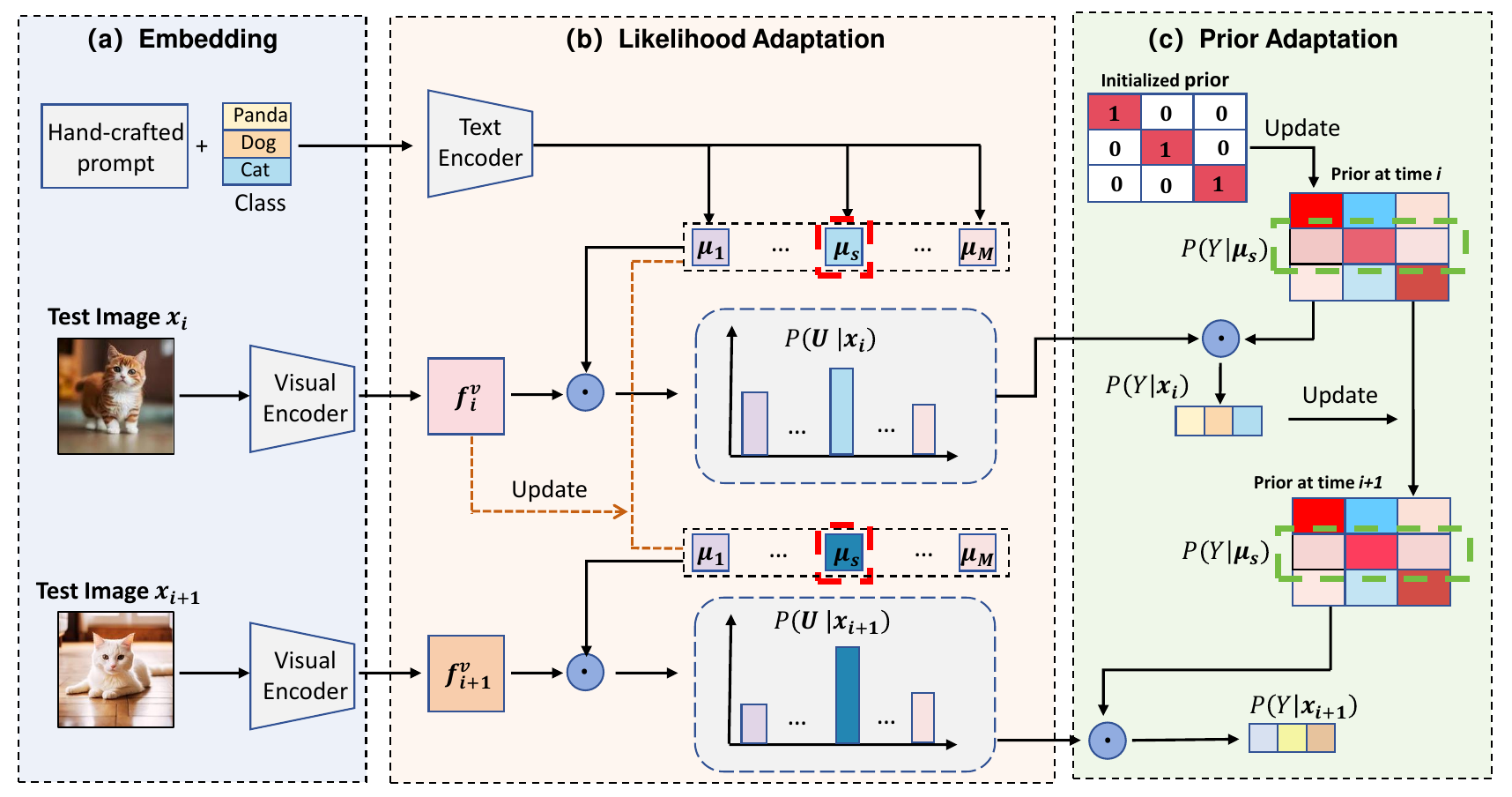}
  \end{center}
     \caption{
    Overview of the proposed \textbf{B}ayesian \textbf{C}lass \textbf{A}daptation (BCA) method. When deploying CLIP to a test environment, $M$ class embeddings are initialized based on hand-crafted prompts, and the prior for each class embedding is initialized as a one-hot vector with the corresponding class set to 1.
    (a) Embedding: when $i$-th image arrives, it is encoded into a visual embedding $\bm{f}^v_i$ using visual encoder. 
    (b) Likelihood adaptation: the probability $P(\bm{U}|\bm{x}_i)$ is calculated based on current likelihood to find class embedding $\bm{\mu}_s$ with the highest probability. This $\bm{\mu}_s$ is then updated using statistical method with $\bm{f}^v_i$ to adapt the likelihood.
    (c) Prior adaptation: the posterior $P(Y|\bm{x}_i)$ is calculated by multiplying $P(\bm{U}|\bm{x}_i)$ by the current prior. And the prior of $s$-th class embedding $P(Y|\bm{\mu}_s)$ is adapted with this posterior.
}
  \label{fig:short}
  \vspace{-0.5cm}
\end{figure*}

\vspace{-0.3cm}
\section{Method}\label{sec:method}



\textbf{Problem Setting.} In this work, we address the challenge of test-time adaptation for a pre-trained visual-language model, CLIP \cite{radford2021learning}. Given a sequence of images $\{\bm{x}_i\}_{i=1}^n$ that arrive sequentially, the model needs to predict the label for each {single} image $\bm{x}_i$ immediately in an online manner. The goal is to improve the classification accuracy of the model on test data $\{\bm{x}_i\}_{i=1}^n$ while preserving inference efficiency.

For a $K$-class classification problem at test time, the CLIP model typically prepares $K$ hand-crafted prompts, such as $\{\texttt{a photo of} [\texttt{Class }k]\}_{k=1}^K$, and sends them to the text encoder $E_t$ to obtain their normalized text embeddings $\bm{F}^t =\{\bm{f}_{k}^t\}_{k=1}^K$. The visual encoder $E_v$ is then used to encode the test images $\bm{x}_i$ to obtain their normalized visual embeddings $\bm{f}_{i}^v$. Finally, the model computes the cosine similarity between $\bm{f}_{i}^v$ and $\bm{F}^t$ to determine the prediction as follows:
\begin{equation}
    P(Y|\bm{x}_i) = \frac{\exp(\cos(\bm{f}_{i}^v, \bm{F}^t))}{\sum_{j=1}^K \exp(\cos(\bm{f}_{i}^v, \bm{f}_{j}^t))},
    \label{Eq:1}
\end{equation}
where $\exp(\cdot)$ denotes the exponential function and $\cos(\cdot, \cdot)$ represents the cosine similarity function.

\subsection{Analysis}\label{Sec:3.1}


In this section, we use Bayes theorem to analyze the process of how the model generates predictions when a sample $\bm{x}_i$ arrives. We consider a more generalized scenario where we have $M$ class embeddings $\{\bm{\mu}_m\}_{m=1}^M$ to perform $K$ category zero-shot classification on the sample $\bm{x}_i$. Here, $M$ can be greater than or equal to $K$. Specifically, $M \geq K$ because some methods \cite{karmanov2024efficient} try to store visual embeddings during the test phase to increase class embeddings. In contrast, many other methods \cite{shu2022test,feng2023diverse} maintain $M=K$, updating each class embedding directly.
The posterior can be formulated as follows:
\begin{equation}
\begin{aligned}
    P(Y|\bm{x}_i) &=\sum_{m=1}^M P(Y,\bm{\mu}_m|\bm{x}_i) \\
    &=\sum_{m=1}^M  P(\bm{\mu}_m|\bm{x}_i) * P(Y|\bm{x}_i,\bm{\mu}_m) \\
    &= \sum_{m=1}^M P(\bm{\mu}_m|\bm{x}_i)* P(Y|\bm{\mu}_m).
    \label{Eq:2}
    \end{aligned}
\end{equation}
The first line is based on the law of total probability. 
The second line uses the definition of conditional probability. The key assumption in the third line is that once we know the class embeddings $\bm{\mu}_m$, the information of $\bm{x}_i$ does not further change the probability of $Y$ \cite{kruschke2010bayesian}.
From the Eq. \eqref{Eq:2}, we can divide the steps of predicting the label of $\bm{x}_i$ into two parts. First, we calculate the probability of $\bm{x}_i$ belonging to each class embedding $\bm{\mu}_m$ and then multiply it with the prior $P(Y|\bm{\mu}_m)$.
Based on Bayes theorem, $P(\bm{\mu}_m|\bm{x}_i)$ can be further written as:
\begin{equation}
\begin{aligned}
    P(\bm{\mu}_m|\bm{x}_i) &= \frac{P(\bm{x}_i|\bm{\mu}_m)*P(\bm{\mu}_m)}{P(\bm{x}_i)} \\
    &=\frac{P(\bm{x}_i|\bm{\mu}_m)*P(\bm{\mu}_m)}{\sum_{j=1}^M P(\bm{x}_i, \bm{\mu}_j)}\\
    &=\frac{P(\bm{x}_i|\bm{\mu}_m)*P(\bm{\mu}_m)}{\sum_{j=1}^M P(\bm{x}_i| \bm{\mu}_j) * P(\bm{\mu}_j)}
    \label{Eq:3}
\end{aligned}
\end{equation}
where $P(\bm{x}_i|\bm{\mu}_m)$ represents likelihood, and $P(\bm{\mu}_m)$ is the probability of each class embedding. Generally, it is assumed that the probability of all class embeddings $P(\bm{\mu}_m)$ is the same, that is, assuming $P(\bm{\mu}_m)=\frac{1}{M}$. 
By substituting Eq. \eqref{Eq:3} and $P(\bm{\mu}_m)=\frac{1}{M}$ back into Eq. \eqref{Eq:2}, we can obtain:
\begin{equation}
    P(Y|\bm{x}_i) = \sum_{m=1}^M \frac{ P(\bm{x}_i|\bm{\mu}_m)}{\sum_{j=1}^M P(\bm{x}_i| \bm{\mu}_j)}* P(Y|\bm{\mu}_m).
    \label{Eq:5}
\end{equation}

\textbf{Remark:} \textbf{1): Example:} For the original CLIP model, the number of class embeddings $M=K$,
the class embeddings are set as $\bm{\mu}_k = \bm{f}_{k}^t$, and the likelihood is defined as $P(\bm{x}_i|\bm{\mu}_k) \propto \exp(\cos(E_v(\bm{x}_i),\bm{\mu}_k)) = \exp(\cos(\bm{f}_{i}^v,\bm{\mu}_k))$. It is important to note that the prior of pre-trained CLIP $P(Y|\bm{\mu}_k) \in R^K$ is a one-hot vector, where the $k$-th element is 1 and all other are 0. This means that when a sample belongs to the $k$-th class embedding, its probability of being in $k$-th category is 1. By substituting these into Eq. \eqref{Eq:5}, we can obtain Eq. \eqref{Eq:1}.
\textbf{2): Core factors:} From the above derivation, it is evident that the core factors affecting posterior are the likelihood $P(\bm{x}|\bm{\mu})$ and prior $P(Y|\bm{\mu})$. Existing methods focus on adapting the likelihood while keeping the prior fixed. In the following sections, we aim to adapt both the likelihood and the prior to improve the model's performance in new environments.

\vspace{-0.2cm}
\subsection{Likelihood Adaptation}


Here, we consider a setting where $M\geq K$, which is more general.
From Eq. \eqref{Eq:5},  it is clear that the accuracy of the final posterior is directly influenced by the likelihood $P(\bm{x}_i|\bm{\mu}_m)$. Recently, various methods have focused on updating this term to make class embedding more suitable for test distribution. For example, TPT \cite{shu2022test}, DiffTPT \cite{feng2023diverse} and C-TPT \cite{yoonc} apply prompt tuning to update text embeddings to update $\bm{\mu}_m$, TDA \cite{karmanov2024efficient} applies feature adapter to add the visual embedding as the new class embeddings, 
DOTA \cite{han2024dota} uses Gaussian distribution $P(\bm{x}_i|\bm{\mu}_m, \bm{\Sigma}_m) = N(\bm{x}_i|\bm{\mu}_m, \bm{\Sigma}_m)$, and continuously update $\bm{\mu}_m$ and $\bm{\Sigma}_m$ based on statistical method.
In this work, we follows original CLIP that assume the likelihood $P(\bm{x}_i|\bm{\mu}_m) \propto \exp(\cos(E_v(\bm{x}_i),\bm{\mu}_m))$, which only requires updating $\bm{\mu}_m$ using statistical method.

Specifically, the text encoder is used to encode $M$ hand-crafted prompts to obtain $M$ text embeddings as the initial class embeddings $\{\bm{\mu}_m\}_{m=1}^M$, where the $m$-th text embedding is obtained by encoding the hand-crafted prompt of the $m\%K$-th category. 
And as the test image $\bm{x}_i$ continues to arrive, the $\bm{\mu}_{m}$ is constantly updated based on $\bm{x}_i$.
Specifically, the visual encoder is used to map the image $\bm{x}_i$ to a visual embedding $\bm{f}_{i}^v$. Then the probability that it belongs to each class embedding $\bm{\mu}_m$ is calculated based on Eq. \eqref{Eq:3}:
\begin{equation}
  \begin{aligned}
P(\bm{\mu}_m|\bm{x}_i) = \frac{P(\bm{x}_i|\bm{\mu}_m)}{\sum_{j=1}^M P(\bm{x}_i| \bm{\mu}_j)}=\frac{\exp(\cos(\bm{f}_{i}^v, \bm{\mu}_m))}{\sum_{j=1}^M \exp(\cos(\bm{f}_{i}^v, \bm{\mu}_j))}
\label{Eq:6}
    \end{aligned}
\end{equation}
Then the class embedding $\bm{\mu}_{s}$ with highest probability is selected. If the probability $P(\bm{\mu}_{s}|\bm{x}_i)$ exceeds the threshold $\tau$, $\bm{\mu}_{s}$ is updated as follows:
\begin{equation}
  \begin{aligned}
    \bm{\mu}_{s} =& Norm(\frac{\bm{C}_1[s]*\bm{\mu}_{s} + \bm{f}_i^v}{\bm{C}_1[s]+1}),\\
    &\bm{C}_1[s] = \bm{C}_1[s]+1,
    \label{Eq:7}
    \end{aligned}
\end{equation}
where $Norm(\cdot)$ denotes vector normalization, and $\bm{C}_1[s]$ is a counter initialized as $\bm{n}_1$ to count samples for the $\bm{\mu}_{s}$.

\subsection{Prior Adaptation}

As mentioned earlier, the prior $P(Y|\bm{\mu}_m)$ in pre-trained CLIP model is a one-hot vector, where the $m\%K$-th element is 1. However, the prior should be different in different environments. For example, consider a person who has a fever. During the COVID-19 period, we naturally think that he is more likely to have COVID-19, whereas in normal period, we might assume he is more likely to have common cold. This illustrates that the prior $P(disease|fever)$ can vary significantly depending on the environment. Therefore, it is crucial to develop a more flexible and adaptive prior that can handle distribution shifts and improve the robustness of the model in real-world applications.

In our method, the prior $P(Y|\bm{\mu}_m)$ is initialized as a $K$-dimensional one-hot vector, in which its $m\%K$-th element is 1 from original CLIP. And the prior will be continuously updated as the samples arrive.
When the $i$-th image $\bm{x}_i$ arrives, the posterior $P(Y|\bm{x}_i)$ is calculated based on the current prior $P(Y|\bm{\mu}_m)$ and the probability $P(\bm{x}_i|\bm{\mu}_m)$, which is calculated in the likelihood adaptation, according to Eq. \eqref{Eq:2} as follows:
\begin{equation}
    P(Y|\bm{x}_i) = \sum_{m=1}^M \frac{\exp(\cos(\bm{f}_{i}^v, \bm{\mu}_m))}{\sum_{j=1}^M \exp(\cos(\bm{f}_{i}^v, \bm{\mu}_j))}* P(Y|\bm{\mu}_m).
    \label{Eq:8}
\end{equation}
where the posterior $P(Y|\bm{x}_i)$ is also the prediction for $\bm{x}_i$.
Then this posterior is also used to update the prior of $s$-th class embedding $P(Y|\bm{\mu}_{s})$, where $\bm{\mu}_{s}$ is the one selected in the likelihood adaptation, as follows:
\begin{equation}
  \begin{aligned}
    P(Y|\bm{\mu}_{s}) =& \frac{\bm{C}_2[s]*P(Y|\bm{\mu}_{s})+P(Y|\bm{x}_i)}{\bm{C}_2[s]+1},\\
    &\bm{C}_2[s] = \bm{C}_2[s]+1,
    \label{Eq:9}
    \end{aligned}
\end{equation}
where $\bm{C}_2[s]$ is a counter initialized as $\bm{n}_2$ to count samples for the $s$-th class embedding.
It is worth noting that there are two different counters $\bm{C}_1$ and $\bm{C}_2$ in our method. The most intuitive understanding is that the learning rates used to update the prior $P(Y|\bm{\mu}_{s})$ and class embedding $\bm{\mu}_{s}$ are different.
By using the $P(Y|\bm{x}_i)$ to update $P(Y|\bm{\mu}_{s})$, the model can continuously adapt to the evolving test distribution, leading to more accurate and robust predictions. 

\begin{algorithm}[t]
  \caption{{{B}ayesian {C}lass {A}daptation}}
  \label{alg:1}
  \textbf{Input}: A pre-trained CLIP model, unlabeled test data $\{(\bm{x}_i)\}_{i=1}^{n}$, $M$ hand-crafted prompts, hyperparameters $\tau$, $\bm{n}_1$ and $\bm{n}_2$\\
  \textbf{Procedure}: 
  \begin{algorithmic}[1] 
  \STATE Initialize the $M$ class embeddings $\bm{U} = [\bm{\mu}_{1}^T;\cdots;\bm{\mu}_{M}^{T}]$ by encoding $M$ hand-crafted prompts with text encoder of pre-trained CLIP;
  \STATE Initialize the prior $\bm{V} = [P(Y|\bm{\mu}_{1}), \cdots, P(Y|\bm{\mu}_{M})]$, where the $m$-th row is a one-hot vector with the $m\%K$-th element being 1;
  \STATE Initialize $\bm{C}_1$ and $\bm{C}_2$ as an $M$-dimensional vector with the values $\bm{n}_1$ and $\bm{n}_2$ respectively;
  \FOR{$i$ = 1:$n$}
   \STATE Map test image $\bm{x}_i$ to visual embedding $\bm{f}^v_i$ ;
  \STATE Calculate the probability $P(\bm{U}|\bm{x}_i)$ of belonging to each class embedding and posterior $P(Y|\bm{x}_i)$ with current $\bm{U}$ and $\bm{V}$ based on Eq. \eqref{Eq:10};
  \STATE Select the most likely class embedding index $s$ based on $P(\bm{U}|\bm{x}_i)$;
  \IF{$P(\bm{U}[s]|\bm{x}_i)>\tau$}
  \STATE {Adapt} $\bm{U}[s]$ with $\bm{f}^v_i$, and $\bm{C}_1[s]$ based on Eq. \eqref{Eq:7};
  \STATE {Adapt} $\bm{V}[s]$ with $P(Y|\bm{x}_i)$, and $\bm{C}_2[s]$ based on Eq. \eqref{Eq:9};
  \ENDIF
  \STATE \textbf{return} posterior $P(Y|\bm{x}_i)$.
  \ENDFOR
  \end{algorithmic}
  \end{algorithm}

\begin{table*}[t]
\begin{small}
\begin{center}
     \caption{Comparison with the state-of-the-art methods on {\it OOD benchmark}. 
     Metric: classification accuracy (\%);
     Bp-free: backpropagation-free at test time;
     Average: mean accuracy across all datasets;
     OOD Average: mean accuracy across four OOD datasets excluding ImageNet.}
     \label{tab:1}
\resizebox{\textwidth}{!}{
\begin{tabular}{l|cc|ccccc|cc}
\hline
\multicolumn{10}{c}{\textbf{Visual Backbone: ResNet-50}}  \\ \hline
Method &Venue &Bp-free  &ImageNet &ImageNet-A &ImageNet-V2 &ImageNet-R &ImageNet-S &Average &OOD Average  \\ \hline

TPT \cite{shu2022test}         & NIPS22  & \XSolidBrush  &60.74 &26.67 &54.70 &59.11 &35.09 &47.26& 43.89 \\
DiffTPT \cite{feng2023diverse}  & ICCV23  & \XSolidBrush  &60.80 &\bf31.06 &55.80 &58.80& 37.10 &48.71 &45.69 \\
C-TPT \cite{yoonc}  & ICLR24  & \XSolidBrush  &61.2   &25.6 &54.8 &59.7 &35.7 &47.4   &44.0 \\
TDA \cite{karmanov2024efficient}  & CVPR24  & \CheckmarkBold  &61.35 &30.29 &55.54 &62.58 &\bf38.12 &49.58 &46.63\\\hline
CLIP \cite{radford2021learning} &ICML21   & \CheckmarkBold   &59.81 &23.24 &52.91 &60.72 &35.48 &46.43 &43.09  \\ 
CLIP with LA & Ours  &\CheckmarkBold   & 60.95 & 28.53 & 55.59  &61.88 & 37.21& 48.83  & 45.80\\ 
CLIP with PA & Ours  &\CheckmarkBold   & 61.15  &28.66 & 55.41  &61.97 &37.41 &48.92   &45.86 \\ 
BCA & Ours  &\CheckmarkBold   & \bf61.81  & 30.35& \bf56.58  &\bf62.89 &38.08 &\bf49.94   &\bf46.98 \\ 
\hline\hline
\multicolumn{10}{c}{\textbf{Visual Backbone: ViT-B/16}}  \\ \hline
 
TPT \cite{shu2022test}         & NIPS22  & \XSolidBrush  &68.98 &54.77 &63.45 &77.06 &47.94 &62.44 &60.81 \\
DiffTPT \cite{feng2023diverse}  & ICCV23  & \XSolidBrush  &\bf70.30 &55.68 &\bf65.10 &75.00 &46.80 &62.28 &60.52 \\
C-TPT \cite{yoonc}  & ICLR24  & \XSolidBrush  & 69.3  & 52.9 &63.4 &78.0 &48.5&62.4 &60.7 \\
TDA \cite{karmanov2024efficient}  & CVPR24  & \CheckmarkBold  & 69.51& 60.11& 64.67& 80.24 &50.54 &65.01 &63.89 \\
MTA \cite{zanella2024test}  &  CVPR24 &  \CheckmarkBold &70.08 &58.06 &64.24 &78.33 &49.61 &64.06 & 62.56\\
PromptAlign \cite{samadhalign} &  NIPS24 & \XSolidBrush & -  &59.37 &65.29 &79.33 &50.23 & -&63.55 \\
\hline
CLIP \cite{radford2021learning} &ICML21   & \CheckmarkBold &  68.34 &49.89&61.88 &77.65 &48.24 &61.20 &59.42  \\ 
CLIP with LA & Ours  &\CheckmarkBold   & 68.99  &58.62 &  63.64 &79.71 &49.58 & 64.11  & 62.89\\ 
CLIP with PA & Ours  &\CheckmarkBold   &  69.16 &59.39 & 64.21  &79.68 & 49.88&64.46   & 63.29\\ 
BCA & Ours  &\CheckmarkBold   & 70.22  & \bf61.14 &  64.90 &\bf80.72 &\bf50.87 &\bf65.37   &\bf64.16 \\ \hline
  \end{tabular}}
\end{center}
\end{small}
\vspace{-0.5cm}
\end{table*}

\subsection{{Our method}}

By integrating the above two points, we first initialize $M$ class embeddings based on hand-crafted prompts, and initialize the prior of each class embedding as a one-hot vector, where the corresponding category is set to 1, that is defining $\bm{U} = [\bm{\mu}_{1}^T;\bm{\mu}_{2}^T;\cdots;\bm{\mu}_{M}^{T}] \in R^{d*M}$, where $d$ is the is the dimension of the embedding and $T$ represents transpose, and $\bm{V} = [P(Y|\bm{\mu}_{1}), P(Y|\bm{\mu}_{2}), \cdots, P(Y|\bm{\mu}_{M})] \in R^{M*K}$. 
In addition, $\bm{C}_1$ and $\bm{C}_2$ are initialized to an $M$-dimensional vector with the values $\bm{n}_1$ and $\bm{n}_2$ respectively to update the class embedding and its prior.
As the test data $\bm{x}_i$ continues to arrive, we first calculate the probability belongs to each class embedding based on Eq. \eqref{Eq:6} and
posterior based on Eq. \eqref{Eq:5}, which can be rewritten in matrix form as follows:
\begin{equation}
\begin{aligned}
    P(\bm{U}|\bm{x}_i) &=  \texttt{Softmax}(\bm{f}_i^v*\bm{U}),\\
    P(Y|\bm{x}_i) &=  \texttt{Softmax}(\bm{f}_i^v*\bm{U})* \bm{V},
    \label{Eq:10}
    \end{aligned}
\end{equation}
where $\texttt{Softmax}(\cdot)$ represents the softmax function.

Then the most likely class embedding $\bm{\mu}_{s}$ is selected based on $P(\bm{U}|\bm{x}_i)$, where $s = \texttt{argmax}_m P(\bm{U}[m]|\bm{x}_i)$. If $P(\bm{U}[s]|\bm{x}_i)>\tau$, 
$\bm{U}[s]$ and $\bm{V}[s]$ are updated by the visual embedding $\bm{f}_i^v$ and the posterior $P(Y|\bm{x}_i)$, which affect the likelihood and prior. These updates are performed using statistical method as shown in Eq. \eqref{Eq:7} and Eq. \eqref{Eq:9} respectively. Our method is summarized in Algorithm \ref{alg:1}.

\vspace{-0.3cm}
\section{Experiments}\label{sec:experiment}
\subsection{Experimental Setup}

\textbf{Benchmarks.}
In our experiments, we evaluated the effectiveness and robustness of our proposed method using two benchmarks: the Cross Domain benchmark and the Out-of-Distribution (OOD) benchmark. These benchmarks have been previously utilized in related research \cite{shu2022test} to test the adaptability of vision-language models during the inference phase. 
The Cross Domain benchmark, on the other hand, evaluates the model's performance across a wide range of image classification tasks from different domains, consisting of ten datasets: Aircraft \cite{maji2013fine}, Caltech101 \cite{fei2004learning}, Cars \cite{krause20133d}, DTD \cite{cimpoi2014describing}, EuroSAT \cite{helber2019eurosat}, Flower102 \cite{nilsback2008automated}, Food101 \cite{bossard2014food}, Pets \cite{parkhi2012cats}, SUN397 \cite{xiao2010sun}, and UCF101 \cite{soomro2012ucf101}. Each dataset represents a unique domain with its own set of classes, allowing us to assess the model's adaptability and generalization capabilities in real-world scenarios where the class spaces may vary significantly.
The OOD benchmark focuses on assessing the model's ability to handle data that differs significantly from the training set, using four specific datasets derived from ImageNet \cite{deng2009imagenet}: ImageNet-A \cite{hendrycks2021natural}, ImageNet-V2 \cite{recht2019imagenet}, ImageNet-R \cite{hendrycks2021many}, and ImageNet-S \cite{wang2019learning}. These datasets are designed to challenge the model's generalization to novel and unseen scenarios, providing insights into its robustness and reliability. 
The OOD benchmark focuses on assessing the model's ability to handle data that differs significantly from the training set, using four specific datasets derived from ImageNet \cite{deng2009imagenet}: ImageNet-A \cite{hendrycks2021natural}, ImageNet-V2 \cite{recht2019imagenet}, ImageNet-R \cite{hendrycks2021many}, and ImageNet-S \cite{wang2019learning}. These datasets are designed to challenge the model's generalization to novel and unseen scenarios, providing insights into its robustness and reliability.

\noindent\textbf{{Implementation details.}}
Our experiment is conducted on the Pytorch platform. Following \cite{shu2022test,feng2023diverse}, we adopt pre-trained CLIP models with ResNet-50 \cite{he2016deep} and ViT-B/16 \cite{dosovitskiy2020image} backbones as the visual encoder and a Transformer \cite{vaswani2017attention} as the text encoder in our experiments. The batch size is set to 1 to suit the application scenario of test time adaptation.
For hyperparameters $\tau/\bm{n}_1/\bm{n}_2$, these set as {$0.3/30000/10$} in OOD benchmark, and set as {$0.35/50000/10$} in Cross Domain benchmark. 
The top-1 accuaracy is used as evaluation metric. All the experiments are conducted on RTX 4070 Ti SUPER GPU.

\vspace{-0.2cm}
\subsection{Comparisons with State-of-the-art}

\begin{table*}[t]
\caption{Comparison with the state-of-the-art methods on {\it Cross Domain benchmark}. 
     Metric: classification accuracy (\%).}
\resizebox{\textwidth}{!}{
\begin{tabular}{l|cccccccccc|c}
\hline
\multicolumn{12}{c}{\textbf{Visual Backbone: ResNet-50}}  \\ \hline
Method   &Aircraft &Caltech101 &Cars &DTD &EuroSAT &Flower102 &Food101 &Pets &SUN397 &UCF101 &Average \\ 
\hline
TPT \cite{shu2022test} &17.58 &87.02 &58.46 &40.84 &28.33 &62.69 &74.88 &84.49 &61.46 &60.82 &57.66  \\
DiffTPT \cite{feng2023diverse}&17.60 &86.89 &60.71 &40.72 &41.04 &63.53 &79.21 &83.40 &62.72 &62.67 &59.85   \\
C-TPT \cite{yoonc} & 17.5 &87.4   & 57.3  &43.1 & 29.4  &65.3   & 76.0  &84.0 & 62.1  & 60.7  & 58.3  \\
TDA \cite{karmanov2024efficient} &17.61 &\bf89.70 &57.78 &43.74 &42.11 &\bf68.74 &77.75 &\bf86.18 &62.53 &\bf64.18 &61.03   \\
HisTPT\cite{zhang2024historical}&18.1  & 87.2  & \bf61.3  & 41.3& \bf42.5  & 67.6  & \bf81.3  &84.9 & 63.5  & 64.1  &61.2   \\ \hline
CLIP \cite{radford2021learning} &16.11 &87.26 &55.89 &40.37 &25.79 &62.77 &74.82 &82.97 &60.85 &59.48 &56.63 \\
CLIP with LA & 17.85 & 88.96 &57.11   &43.55 &  35.50 &65.57   &76.94   &84.24 & 62.54  & 60.71  &59.30 \\ 
CLIP with PA & 18.10 & 89.08  &  57.56 & 46.06&  38.53&65.02  &76.92   &85.28 & 62.20  &60.45   &59.92 \\ 
BCA & \bf19.89 & \bf89.70  & 58.13  & \bf48.58 & 42.12  & 66.30 &77.19   &85.58  &\bf63.38   &63.51   & \bf61.44\\ \hline
\hline 
\multicolumn{12}{c}{\textbf{Visual Backbone: ViT-B/16}}  \\ \hline
TPT \cite{shu2022test} &24.78 &94.16 &66.87 &47.75 &42.44 &68.98 &84.67 &87.79 &65.50 &68.04 &65.10 \\
DiffTPT \cite{feng2023diverse}       &25.60 &92.49 &67.01 &47.00 &43.13 &70.10 &87.23 &88.22 &65.74 &62.67 &65.47\\
C-TPT \cite{yoonc} &23.9   & 94.1  & 66.7  & 46.8 &48.7   &69.9   &84.5   &87.4 & 66.0  &66.7   &    65.5\\
TDA \cite{karmanov2024efficient} &23.91 &94.24 &67.28 &47.40 &\bf58.00 &71.42 &86.14 &88.63 &67.62 &\bf70.66 &67.53\\
MTA \cite{zanella2024test} &25.20   &94.21   &68.47   &45.90  &45.36   &68.06   &85.00   &88.24 &66.67   &68.69   &   65.58\\
PromptAlign \cite{samadhalign}& 24.80  & 94.01  & 68.50  & 47.24 & 47.86  &  72.39 & \bf86.65  & \bf90.76 & 67.54  & 69.47  &   66.92\\
HisTPT\cite{zhang2024historical} (NIPS24)& 26.9  & 94.5  & \bf69.2  & 48.9 & 49.7  &  71.2 &  89.3 & 89.1 &  67.2 &  70.1 &   67.6\\ 
Zero \cite{farina2024frustratingly} (NIPS24)& 24.42  & 94.14  &  68.48 & 45.86 & 43.77  & 66.82  & 84.58  &87.20 & 66.90  & 68.57  &   65.07\\ \hline
CLIP \cite{radford2021learning}  &23.22 &93.55 &66.11 &45.04 &50.42 &66.99 &82.86 &86.92 &65.63 &65.16 &64.59 \\ 
CLIP with LA &26.85  &93.50   &65.34   &51.89 & 50.82  &71.41   &85.60   &89.99 & 67.09  & 66.87  &66.94 \\ 
CLIP with PA &26.70  & 94.19  & 65.63  & 52.60 & 55.17  & 72.71  & 85.75  &90.30 & 68.04  & 67.01  & 67.81\\ 
BCA &\bf28.59  &\bf94.69   &66.86   & \bf53.49& 56.63  &\bf73.12   &85.97   &90.43 &\bf68.41   &67.59   & \bf68.59\\ \hline
\end{tabular}}
\label{tab:2}
\vspace{-0.5cm}
\end{table*}

For fair comparisons with other method, we set $M=K$
in this section, and follow \cite{karmanov2024efficient} to use multiple context prompt templates for prompt ensembling and generate $K$ initialized class embeddings for image classification. And we compare our method with the following two {major categories of methods}: 
(1) Backpropagation-based methods: these methods require backpropagation during the test time, such as TPT \cite{shu2022test}, DiffTPT \cite{feng2023diverse}, C-TPT \cite{yoonc}, PromptAlign \cite{samadhalign} and HisTPT\cite{zhang2024historical}.
(2) Backpropagation-free methods: these methods do not require backpropagation and are designed to be computationally efficient, such as TDA \cite{karmanov2024efficient}, MTA \cite{zanella2024test} and Zero \cite{farina2024frustratingly}.
In addition, we also report our method in detail, where CLIP \cite{radford2021learning} represents our baseline method; CLIP with LA represents the addition of Likelihood adaptation on the basis of CLIP; CLIP with PA represents the addition of prior adaptation on the basis of CLIP; BCA represents the complete algorithm.

\noindent\textbf{Results on the OOD benchmark.} We first compare BCA with state-of-the-art methods over the OOD benchmark. Table \ref{tab:1} presents the experimental results, highlighting the superior performance of the proposed BCA compared to other methods across various OOD datasets derived from ImageNet.
Specifically, for the ResNet-50 backbone, BCA outperforms TDA by an average of 0.36 in overall accuracy and 0.35 in OOD accuracy. For instance, BCA achieves 61.81 accuracy on the ImageNet dataset, a 0.46 improvement over TDA's 61.35 and a 0.61 improvement over C-TPT's 61.2. On the ImageNet-V2 dataset, BCA reaches 56.58 accuracy, a 1.04 improvement over TDA's 55.54 and a 0.78 improvement over DiffTPT's 55.80. On the ImageNet-R dataset, BCA attains 62.89 accuracy, a 0.31 improvement over TDA's 62.58 and a 3.19 improvement over C-TPT's 59.7. For the ViT-B/16 backbone, BCA continues to exhibit strong performance. On the ImageNet-R dataset, BCA achieves 80.72 accuracy, a 0.48 improvement over TDA's 80.24 and a 1.39 improvement over PromptAlign's 79.33. On the ImageNet-S dataset, BCA reaches 50.87 accuracy, a 0.33 improvement over TDA's 50.54 and a 0.64 improvement over PromptAlign's 50.23. Overall, BCA achieves an average accuracy of 65.37 and an OOD average accuracy of 64.16, representing improvements of 0.36 and 0.27 over TDA's 65.01 and 63.89, respectively.

\noindent\textbf{Results on the Cross Domain benchmark.} 
We also compare BCA with state-of-the-art methods over the cross-domain benchmark. Table \ref{tab:2} presents the experimental results, highlighting the superior performance of the proposed BCA across various cross-domain datasets.
Specifically, for the ResNet-50 backbone, BCA achieves 19.89 accuracy on the Aircraft dataset, a 2.28 improvement over TDA's 17.61 and a 1.79 improvement over HisTPT's 18.1. On the DTD dataset, BCA attains 48.58 accuracy, a 4.84 improvement over TDA's 43.74 and a 5.48 improvement over C-TPT's 43.1. Overall, BCA achieves an average accuracy of 61.44, representing a 0.41 improvement over TDA's 61.03 and a 0.24 improvement over HisTPT's 61.2.
For the ViT-B/16 backbone, BCA continues to exhibit strong performance. On the Aircraft dataset, BCA achieves 28.59 accuracy, a 4.68 improvement over TDA's 23.91 and a 1.69 improvement over HisTPT's 26.9. On the DTD dataset, BCA attains 53.49 accuracy, a 6.09 improvement over TDA's 47.40 and a 4.59 improvement over HisTPT's 48.9. On the Flower102 dataset, BCA reaches 73.12 accuracy, a 1.70 improvement over TDA's 71.42 and a 0.72 improvement over HisTPT's 72.4. Overall, BCA achieves an average accuracy of 68.59, representing a 1.06 improvement over TDA's 67.53 and a 0.99 improvement over HisTPT's 67.6.

\begin{figure*}[t]
\centering
\includegraphics[width=1\linewidth]{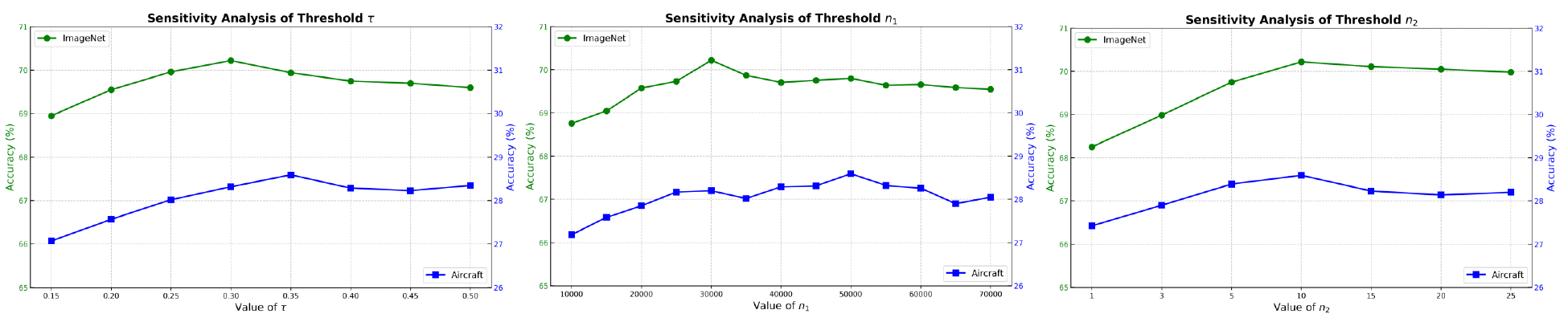}
\caption{Sensitivity analysis with respect to $\tau/\bm{n}_1/\bm{n}_2$ on ImageNet for OOD benchmark and  Aircraft for Cross Domain benchmark using ViT-B/16 as the visual backbone.}
\label{fig:3}
\vspace{-0.5cm}
\end{figure*}

\subsection{Ablation Studies}

\noindent{\bf Component analysis.} 
In this experiment, we validated the effectiveness of each component of the proposed method on two benchmarks as shown in Table \ref{tab:1} and Table \ref{tab:2}. Based on the experimental results, we have summarized the following patterns: 1) Adapting likelihood and prior can both enable the CLIP model to better adapt to the test environment: By comparing the original CLIP model with the CLIP models that incorporate LA and PA, we found that CLIP with LA and CLIP with PA achieved significant improvements in classification accuracy across all datasets. This demonstrates that adapting the pre-trained CLIP model to the test environment can effectively improve the model's accuracy, thereby validating the correctness of Analysis in Section \ref{Sec:3.1}. 2) Adapting likelihood and prior can work synergistically: By comparing BCA with CLIP with LA and BCA with CLIP with PA, it is evident that BCA achieved even further performance enhancements, indicating that there is a complementary effect between LA and PA, allowing them to combine to enhance the model's generalization ability. 3) Prior adaptation is equally crucial: By comparing CLIP with PA to the original CLIP and BCA to CLIP with LA, we found that adding prior adaptation could significantly boost the model's accuracy, which is often ignored in previous model designs, highlighting the importance of domain-specific prior knowledge in improving the model's cross-domain adaptability.

\begin{table}[h]
\begin{small}
\begin{center}
     \caption{Comparison with the state-of-the-art methods on {\it ImageNet} in terms of efficiency (Testing Time, Memory usage) and effectiveness (Accuracy). 
     Visual Backbone: ResNet-50;
     GPU: RTX 4070 Ti SUPER GPU.}
     \label{tab:3}
\begin{tabular}{lcccc}
\hline
Method  &Testing time &Memory usage &Accuracy &Gain  \\ \hline
CLIP    &2.23min &753M &59.81 &0 \\ \hline
TPT &572.13min &21396M & 60.74&+0.93\\ 
TDA&11.93min & 1174M& 61.35&+1.54\\
BCA &2.42min &757M &61.81 &+2.00 \\ 
\hline
  \end{tabular}
\end{center}
\end{small}
\end{table}

\noindent{\bf Inference time comparison.}
In this experiment, we evaluated the efficiency and effectiveness of the proposed BCA method on the ImageNet dataset using ResNet-50 as the visual backbone on RTX 4070 Ti SUPER GPU. 
We compared our BCA with TPT and TDA, the experimental results are shown in Table \ref{tab:3}. From the table, we can see that the BCA method outperforms other methods in several aspects:
1) Inference Time: The inference time of the BCA method is significantly reduced compared to both methods that require gradient backpropagation and those that do not. For example, TPT takes 572.13 minutes, while BCA only takes 2.42 minutes. Even compared to TDA, which does not require gradient backpropagation, BCA is still much faster, taking only 2.42 minutes compared to TDA's 11.93 minutes.
2) Memory Usage: The memory usage of the BCA method is also significantly lower than that of other methods, especially those that require gradient backpropagation. For example, TPT uses 21396M of memory, while BCA uses only 757M. Compared to TDA, which uses 1174M, BCA still uses less memory, which is on par with the baseline CLIP.
3) Classification Accuracy: Despite its excellent performance in memory usage and inference time, BCA also achieves higher classification accuracy. For example, BCA achieves an accuracy of 61.81, which is a 2.00 improvement over the baseline CLIP. This is slightly better than TDA, which achieves an accuracy of 61.35.

\noindent{\bf Parameter sensitivity analysis.}
In this experiment, we investigated the settings of hyperparameters $\tau/\bm{n}_1/\bm{n}_2$ to analyze the sensitivity of our method. We conducted experiments on the ImageNet for OOD benchmark and Aircraft for Cross Domain benchmark using ViT-B/16 as the visual backbone. The experimental results are shown in Figure \ref{fig:3}. It can be found that the accuracy first increases and then decreases with the increase of the three hyperparameter values.
And our method is also robust to the hyperparameters $\tau/\bm{n}_1/\bm{n}_2$. This suggests that our method maintains good performance under various hyperparameter settings, making it suitable for different tasks and datasets.

\section{Conclusion}\label{sec:conclusion}

In this work, we propose {B}ayesian {C}lass {A}daptation (BCA). Unlike previous methods that focus solely on likelihood estimation, BCA incorporates the continuous updating of the prior distribution, which is crucial for handling distribution shifts in real-world applications. By employing a statistical approach to update both the likelihood and the prior, BCA enhances the robustness and accuracy of zero-shot image classification, particularly when the test data distribution differs significantly from the pre-trained data. Our experiments demonstrate that BCA not only outperforms existing methods on Out-of-Distribution (OOD) and Cross Domain tasks but also maintains high inference speed and efficiency, making it a effective solution for real-world scenarios.

\section*{Acknowledgement}
This work was supported in part by Chinese National Natural Science Foundation Projects U23B2054, 62276254, 62306313, the Beijing Science and Technology Plan Project Z231100005923033, Beijing Natural Science Foundation L221013, and InnoHK program.

{
    \small
    \bibliographystyle{ieeenat_fullname}
    \bibliography{main}
}
\clearpage
\setcounter{page}{1}
\maketitlesupplementary

\section{Algorithm complexity analysis}
In this section, we analyze and compare the complexity of our method BCA with the baseline method CLIP. 
Assume that BCA uses $M$ class embeddings $\bm{U} = [\bm{\mu}_{1}^T;\bm{\mu}_{2}^T;\cdots;\bm{\mu}_{M}^{T}] \in R^{d*M}$, where $d$ is the is the dimension of the embedding and $T$ represents transpose, and $\bm{V} = [P(Y|\bm{\mu}_{1}), P(Y|\bm{\mu}_{2}), \cdots, P(Y|\bm{\mu}_{M})] \in R^{M*K}$ for test time adaptation. 
When a sample $\bm{x}_i$ arrives, BCA needs to complete the following steps to get the prediction for the sample $\bm{x}_i$:

\begin{enumerate}
\item Map sample $\bm{x}_i$ to the visual embedding $\bm{f}_i^v$.
\item Calculate the probability of belonging to each class embedding  $P(\bm{U}|\bm{x}_i) =  \texttt{Softmax}(\bm{f}_i^v*\bm{U})$.
\item Compute the final prediction by integrating the prior   $P(Y|\bm{x}_i)=  P(\bm{U}|\bm{x}_i)* \bm{V}$.
\item Update $\bm{U}[s]$ and $\bm{V}[s]$ when $P(\bm{U}[s]|\bm{x}_i)>\tau$, where $s = \texttt{argmax}_m P(\bm{U}[m]|\bm{x}_i)$.
\end{enumerate}
For the first step, the required time $t_1$ is mainly affected by the visual backbone.
For the second step, the required time $t_2$ can be divided into the calculation of $\bm{f}_i^v*\bm{U}$, with a complexity of $\bm{O}(d*M)$, and the softmax operation, with a complexity of $\bm{O}(M)$, so the overall complexity is $\bm{O}(d*M)$.
For the third step, the required time $t_3$ is used to calculate of $P(\bm{U}|\bm{x}_i)* \bm{V}$, and its complexity is $\bm{O}(K*M)$.
Compared with the first three steps, the the required time $t_4$ for the fourth step involves not only the completion of the calculation $t_{4-cal}$, but also the reading and writing of memory $t_{4-rw}$.
The first part $t_{4-cal}$ is used to calculate updated class embeddings and priors, which 
is first used to determine whether it exceeds the threshold $\tau$, with a complexity of $\bm{O}(M)$, and then used to update $\bm{U}[s]$ and $\bm{V}[s]$, with a complexity of $\bm{O}(d+K)$. 
Please note that the update in the fourth step will only be executed if the required conditions are met, so its complexity is sometimes only $\bm{O}(M)$.

\begin{table}[h]
\begin{small}
\begin{center}
\caption{Analysis of the required time for each step on {\it Cross Domain} benchmark.}
\label{tab:time}
{
\begin{tabular}{lccccc}
\hline
  &$t_1$ &$t_2$ &$t_3$ &$t_{4-cal}$ &$t_{4-rw}$  \\ \hline

BCA-RN50 &217.50s&4.11s&0.64s&10.22s&4.19s\\ 
BCA-ViT-B/16 &208.86s&3.18s&0.63s&10.25s&4.32s\\ 
\hline
  \end{tabular}}
\end{center}
\end{small}
\end{table}

Since current deep networks usually require a lot of matrix operations, the first step $t_1$ usually takes the longest time.
Generally speaking, $d>K$, which means $\bm{O}(d*M)>\bm{O}(K*M)$, that is, the time required for the second step $t_2$ is also greater than that for the third step $t_3$. 
In the fourth step, the update is just some very simple operations, so it takes the shortest time $t_{4-cal}$. Therefore, theoretically, $t_1>t_2>t_3>t_{4-cal}$.
To verify this, we verify it on the Cross Domain benchmark, as shown in the Table \ref{tab:time}. The results show that while $t_1$ remains the longest step, consistent with our theoretical analysis, the actual running time for $t_{4-cal}$ is significantly longer than both $t_2$ and $t_3$. Specifically, for both BCA-RN50 and BCA-ViT-B/16, $t_{4-cal}$ takes approximately 10 seconds, which is much longer than $t_3$ (around 0.64s for BCA-RN50 and 0.63s for BCA-ViT-B/16) and even longer than $t_2$ (4.11s for BCA-RN50 and 3.18s for BCA-ViT-B/16). This discrepancy can be attributed to the following reasons:
\begin{enumerate}
\item {\bf Serial Operations}: The update operations in $t_4-cal$ involve multiple serial steps, such as condition checks and sequential updates of multiple variables. These serial operations cannot be fully parallelized, leading to increased time overhead.
In contrast, $t_2$ and $t_3$ primarily involve matrix operations, which can be efficiently parallelized on the GPU. Matrix operations are highly optimized in modern deep learning frameworks, allowing them to run much faster despite their higher theoretical complexity.

\item {\bf Kernel Launch Overhead}: Frequent kernel launches and synchronization operations in $t_4$ can add significant overhead. While $t_2$ and $t_3$ do not involve read and write operations to external variables; they operate on local variables, further reducing memory access overhead.

\end{enumerate}

\begin{figure*}[t]
\centering
\includegraphics[width=1\linewidth]{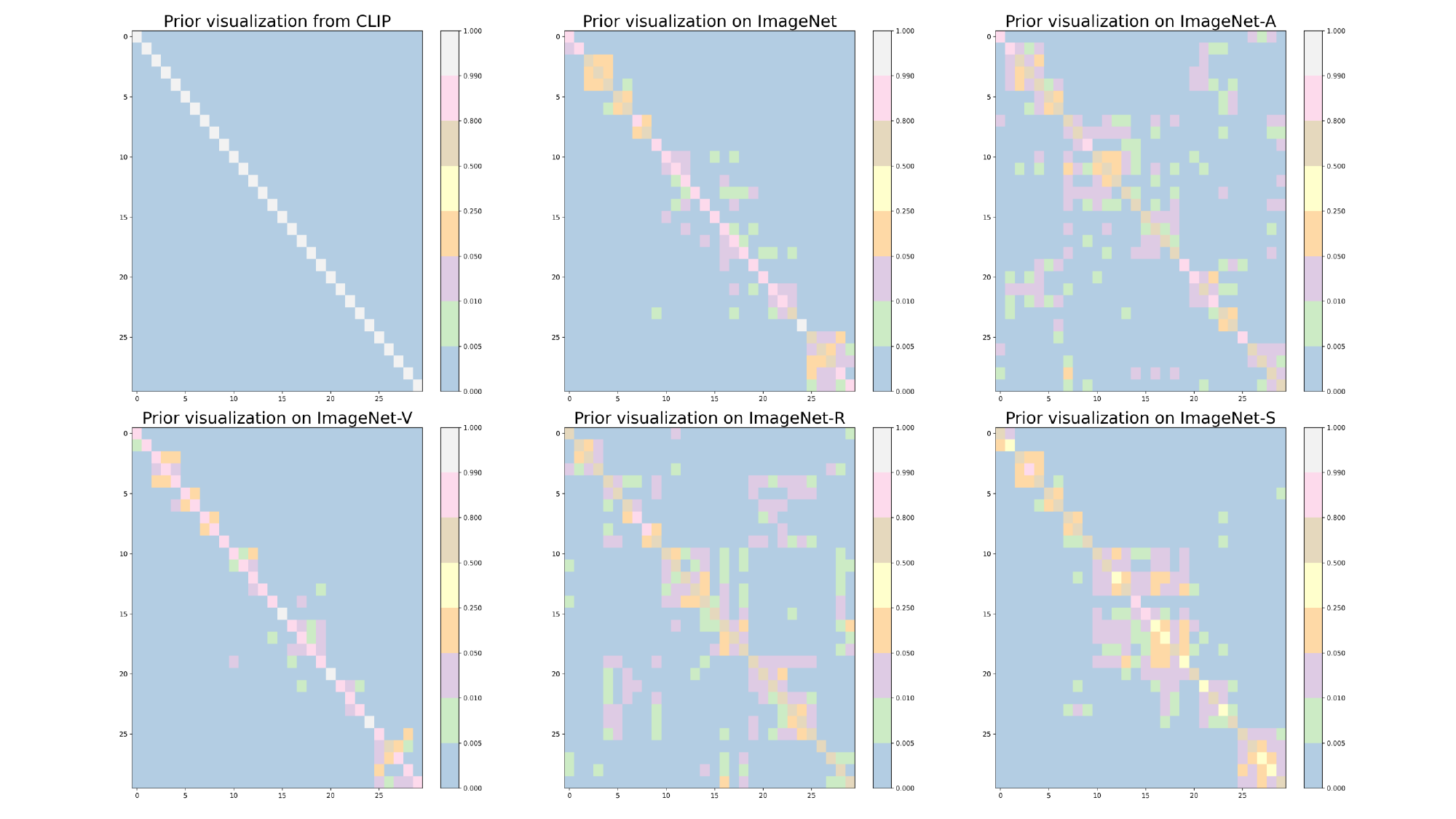}
\caption{Prior visualization on OOD benchmark.}
\label{fig:prior}
\end{figure*}

Therefore, while the theoretical complexity suggests $t_{4-cal}$ should be the shortest, practical considerations such as serial operations and kernel launch overhead, can significantly impact the actual running time. These factors explain why $t_{4-cal}$ is observed to be longer than both $t_2$ and  $t_3$ in practice.

Compared with CLIP, which only has the first two steps and $M=K$, BCA further includes the third and fourth steps. The total time required for the third and fourth steps is less than 10\% of the entire method. This also shows that our method is very efficient.

Furthermore, we analyzed the memory usage  of BCA compared to CLIP. When $M=K$, BCA only requires additional storage $\bm{V}$ on the basis of CLIP. Essentially, this means storing an additional $K*K$ elements. For example, on Imagenet, which has 1000 categories, we only need to store an additional $1000*1000$ matrix, which amounts to $1000*1000*4bytes \approx 4 M$ of extra memory. This additional memory requirement is minimal compared to the overall model size.

\section{More Ablation Studies}

\noindent{\bf Prior visualization.} 
In this experiment, we visualize the prior of the model after performing TTA on ImageNet and its four variant datasets, and compared them with the fixed prior of the CLIP model. Considering that ImageNet contains a large number of categories, we selected the top 30 categories for visualization, as shown in Figure \ref{fig:prior}. Through this experiment, it can be observed that even when classifying the same categories, distribution discrepancies can lead to significant differences in the model's prior. This highlights the importance of dataset characteristics and distributions on model learning and prediction, and underscores the importance of performing prior adaptation.

\begin{table}[h]
\begin{small}
\begin{center}
\caption{Analysis of the number of hand-crafted prompts $M$ on {\it ImageNet}.}
\label{tab:M}
\resizebox{0.45\textwidth}{!}{
\begin{tabular}{lcccc}
\hline
Method  &$M=K$ &$M=2K$ &$M=3K$ &$M=4K$   \\ \hline
CLIP-RN50  &58.15 &57.15 & 57.71 &58.95\\ 
BCA-RN50 &60.54 &60.75 & 61.12 &61.33\\ \hline
CLIP-ViT-B/16  &66.74&66.18&67.39&67.48\\ 
BCA-ViT-B/16 &69.24&69.63&69.95&70.05\\ 
\hline
  \end{tabular}}
\end{center}
\end{small}
\end{table}
\noindent{\bf Analysis of the number of hand-crafted prompts $M$.}
In this experiment, we investigate the impact of using different numbers of hand-crafted prompts on the experimental results, as described in the method section. Notably, in this experiment, we did not perform prompt ensemble as in Section 4.2. The experimental results are shown in Table \ref{tab:M}. Specifically, 
when $M=k$, our hand-crafted template is $\{\texttt{a photo of a } [\texttt{Class }k]\}_{k=1}^K$. 
When $M=2k$, our hand-crafted template is 
$\{\texttt{a photo}$ $\texttt{of a}$ $[\texttt{Class }k]\}_{k=1}^K$
+
$\{\texttt{a origami}$ $[\texttt{Class }k]\}_{k=1}^K$.
When $M = 3k$, our hand-crafted template is 
$\{\texttt{a photo of a}$ $[\texttt{Class }k]\}_{k=1}^K$ 
+ 
$\{\texttt{a origami}$ $[\texttt{Class }k]\}_{k=1}^K$
+
$\{\texttt{art of the}$ $[\texttt{Class }k]\}_{k=1}^K$.
When $M = 4k$, our hand-crafted template is 
$\{\texttt{a photo}$ $\texttt{of a}$ $[\texttt{Class }k]\}_{k=1}^K$
+ 
$\{\texttt{a origami}$ $[\texttt{Class }k]\}_{k=1}^K$
+
$\{\texttt{art of the}$ $[\texttt{Class }k]\}_{k=1}^K$
+ 
$\{\texttt{itap of a}$ $[\texttt{Class }k]\}_{k=1}^K$.
From the experimental results, we can observe that as $M$ increases, the performance of BCA continues to improve, while this is not the case for CLIP \cite{radford2021learning}. The reason is that CLIP does not update class embedding to perform likelihood adaptation and prior adaptation in the new environment, so it is greatly affected by the hand-crafted prompts, while BCA is relatively less affected.

\begin{table}[h]
\begin{small}
\begin{center}
\caption{Prior Adaptaion Analysis Combined with TDA. ImageNet: accuacy on ImageNet dataset. OOD Average: mean accuracy across four Out-of-Distribution datasets. CD Average: mean accuracy across ten Cross Domain datasets.}
\label{tab:tda}
\begin{tabular}{lccc}
\hline
Method   &ImageNet &OOD Average &CD Average   \\ \hline
TDA-RN50  &61.35&46.63&61.03 \\ 
TDA+PA &62.09&47.52&62.23 \\ \hline
TDA-ViT-B/16 &69.51&63.89&67.53\\ 
TDA+PA &70.19&64.58&68.36\\ 
\hline
\end{tabular}
\end{center}
\end{small}
\end{table}
\noindent{\bf Prior Adaptaion Analysis Combined with TDA.} 
In this work, our core strategy is the introduction of prior adaptation. To validate its generalization capability combined with other likelihood adaptation, we integrate this strategy with the existing likelihood adaptation method TDA \cite{karmanov2024efficient}, which optimizes the model's likelihood adaptation by continuously adding visual embeddings as new class embeddings.
The experimental results are shown in Table \ref{tab:tda}. These results demonstrate that prior adaptation effectively enhances the generalization and adaptability of the TDA method, leading to improved performance across multiple datasets and evaluation metrics. This also proves prior adaptation has a good ability to integrate with other likelihood adaptation.

\begin{table}[h]
\begin{small}
\begin{center}
\caption{Performance Comparison for the last 50\% samples. Visual backbone: ViT-B/16. I: ImageNet; A: ImageNet-A; V: ImageNet-V2; R: ImageNet-R; S: ImageNet-S. }
\label{tab:l50}
\begin{tabular}{lccccc}
\hline
Method   &I &A &V&R&S   \\ \hline
CLIP  &67.89 &49.78 &61.36 &77.27 &48.44\\ 
CLIP+LA  &69.17 &59.06 &63.75 &80.08 &50.07\\ 
 CLIP+PA  &69.34 &59.75 &64.11 &79.78 &50.28\\ 
  BCA  &\bf70.68 &\bf61.99 &\bf65.48 &\bf81.14 &\bf51.36\\ \hline
\end{tabular}
\end{center}
\end{small}
\end{table}
\noindent{\bf{Performance Comparison for the last 50\% samples.}} 
In this experiment, we aim to evaluate the long-term performance and adaptability of different methods by focusing on the last 50\% of samples in OOD benchmark. This setup helps us understand how well these methods can adapt to new environments over time. We compared our method, BCA, with CLIP on these later samples.
The results show that BCA consistently outperformed CLIP, achieving the best performance across the last 50\% of samples. This indicates that BCA not only maintains its initial effectiveness but also demonstrates strong adaptability to new and evolving environments. This robust performance over time highlights BCA's potential for real-world applications where data distributions can change dynamically.
In conclusion, the experimental results confirm that BCA is highly effective in adapting to new conditions, making it a promising approach for long-term deployment in vision-language modeling tasks.

\begin{table}[h]
\begin{small}
\begin{center}
\caption{Performance Comparison on OOD benchmark. Visual backbone: ViT-L/14. I: ImageNet; A: ImageNet-A; V: ImageNet-V2; R: ImageNet-R; S: ImageNet-S. }
\label{tab:l14}
\begin{tabular}{lccccc}
\hline
Method   &I &A &V&R&S   \\ \hline
CLIP-ViT-L/14  &74.04 &53.88 &67.69 &87.42 &63.18\\
TDA&76.28 &61.27 &68.42 &\bf88.41 &64.67\\
BCA &\bf77.09& \bf61.62 &\bf69.93 &88.27 &\bf65.41\\  \hline
\end{tabular}
\end{center}
\end{small}
\end{table}

\noindent{\bf {Performance Comparisons on Larger-Scale VLMs.} }
In this experiment, we use ViT-L/14 as the visual backbone to evaluate the performance of our proposed method, BCA, on OOD benchmark. The goal was to assess BCA's effectiveness with a lager visual backbone, which is a common requirement in real-world applications where handling complex and diverse data is essential.
The experimental results, shown in Table \ref{tab:l14}, demonstrate that BCA consistently outperformed other state-of-the-art methods, such as CLIP and TDA, across four of the five tested datasets. Specifically, BCA achieved the highest accuracy on ImageNet (I), ImageNet-A (A), ImageNet-V2 (V), and ImageNet-S (S). While TDA slightly outperformed BCA on ImageNet-R (R) with an accuracy of 88.41\%, BCA still maintained a high accuracy of 88.27\%.
These results highlight BCA's robustness and adaptability in handling OOD data. The consistent performance gains across multiple datasets indicate that BCA can effectively leverage the rich feature representations provided by large-scale pre-trained models like ViT-L/14. This is particularly important as the use of such models becomes increasingly prevalent in both research and industry.

\begin{table}[h]
\begin{small}
\begin{center}
\caption{Ablation Study on Update Strategies for BCA. Visual backbone: ViT-B/16. Dataset: ImageNet.}
\label{tab:ablation_update}
\begin{tabular}{lc}
\hline
Update Strategy   & Accuracy (\%) \\ \hline
Count-based       & 70.22         \\
Momentum-based    & 70.08         \\
Decay-based       & 69.92         \\ \hline
\end{tabular}
\end{center}
\end{small}
\end{table}

\noindent{\bf {Ablation Study on Update Strategies.} }
We evaluated BCA’s robustness across update strategies using ViT-B/16 on ImageNet. Table \ref{tab:ablation_update} shows that Count-based (70.22\%), Momentum-based (70.08\%), and Decay-based (69.92\%) strategies yield comparable accuracies, with a variance of less than 0.3\%. This consistency highlights BCA’s robustness to different update mechanisms, reinforcing the effectiveness of its core design.


\end{document}